\documentclass[10pt, conference, letterpaper]{IEEEtran}
\IEEEoverridecommandlockouts
\usepackage{cite}
\usepackage{amsmath,amssymb,amsfonts}
\usepackage{algorithmic}
\usepackage{graphicx}
\usepackage{textcomp}
\usepackage{xcolor}
\usepackage{caption}

\usepackage[flushleft]{threeparttable}
\usepackage{multirow}
\usepackage{booktabs,caption}

\def\BibTeX{{\rm B\kern-.05em{\sc i\kern-.025em b}\kern-.08em
    T\kern-.1667em\lower.7ex\hbox{E}\kern-.125emX}}

\begin{document}
%%%%%%%%%%%%%%%%%%%%%%%%%%%%%%%%%%%%%%%%%%%%%%%%%%%%%%%%%%
\title{Smart City Intersections: Intelligence Nodes for Future Metropolises}

\author{Zoran~Kosti\'{c}*, Alex~Angus*, Zhengye~Yang**, Zhuoxu~Duan*, \\ Ivan~Seskar***, Gil~Zussman*, Dipankar~Raychaudhuri***
\\
*Dept. of Electrical Engineering, Columbia University, New York City
\\
**Rensselaer Polytechnic Institute
\\
***Winlab, Rutgers
}

\maketitle

%%%%%%%%%%%%%%%%%%%%%%%%%%%%%%%%%%%%%%%%%%%%%%%%%%%%%%%%%%%%%%%%%%%%%%%%%%%%%%
\begin{abstract}
Traffic intersections are the most suitable locations for the deployment of computing, communications, and intelligence services for smart cities of the future. The abundance of data to be collected and processed, in combination with privacy and security concerns, motivates the use of the edge-computing paradigm which aligns well with physical intersections in metropolises.
This paper focuses on high-bandwidth, low-latency applications, and in that context it describes: (i) system design considerations for smart city intersection intelligence nodes; (ii) key technological components including sensors, networking, edge computing, low latency design, and AI-based intelligence; and (iii) applications such as privacy preservation, cloud-connected vehicles, a real-time "radar-screen", traffic management, and monitoring of pedestrian behavior during pandemics. The results of the experimental studies performed on the COSMOS testbed located in New York City are illustrated. Future challenges in designing human-centered smart city intersections are summarized.  

\end{abstract}

%%%%%%%%%%%%%%%%%%%%%%%%%%%%%%%%%%%%%%%%%%%%%%%%%%%%%%%%%%%%%%%%%%%%%%%%%%%%%%
\section{Introduction}

\textbf{Smart cities} should be built with the primary goal of providing social good as defined by local communities~\cite{Snchez2014SmartSantanderIE, smartcities3030052}.
Contemporary technologies provide a plethora of components to support human-centered design of future metropolises. Issues of privacy, security, and local data governance on one hand, and optimization of bandwidth, computational resources, and latency on the other hand, implicate traffic intersections as the best locations for smart city intelligence nodes.

\textbf{Smart-city intersections} are the key locations for emerging smart cities, since city dynamics can be supported by the interconnection and collaboration between neighboring intersection intelligence nodes. The nodes will be equipped with  artificial intelligence (AI)-enabled edge-computing~\cite{10.1145/3523230.3523235} and communications equipment to facilitate automated low-latency data harvesting, inference, and decision making. This will enable the development of technologies like cloud connected vehicles, vehicle to infrastructure communications, and advanced sensory-based tools for alerting pedestrians and assisting handicapped individuals. Future applications will require intense AI-enabled computation, very high communication bandwidths, and ultra-low latencies. 

%%%%%%%%%%%%%%%%%%%%%%%%%%%%%%%%%%%%%%%%%%%%%%%%%%%%%%%%%%%
%This paper reports 
We report the results of research on low-latency real-time applications for smart city intersections in metropolises and architectures,  components, and methods for building intelligent intersection nodes. The research utilizes COSMOS, an experimental testbed located in New York City.

\section{Smart City Intersections}

%%%%%%%%%%%%%%%%%%%%%%%%%%%%%%%%%%%%%%%%%%%%%%%%%%%
%\textbf{The experiments are focused} 
%\textbf{The focus of this paper} 
The focus of this paper is low-latency high-bandwidth applications for smart city intersections. We explore how to support privacy-preserving real-time applications such as collaborative control of cloud connected vehicles and active pedestrian alert and assistance; 
%for exploring how to support real-time applications such as collaborative control of cloud connected vehicles and active pedestrian alert and assistance. 
%This is exemplified by the use of  high-resolution video recording feeds. 
both require the use of a number of sensors including multiple high-resolution video cameras. 
One of the key tasks for video-based applications is to detect and track objects in an intersection with high accuracy. 
%Additionally, the studies are  aimed at exploring 
We explore methods to achieve real time in smart city intersection applications defined by end-to-end latencies under $33.3$ milliseconds. This includes (i) sensor data acquisition; (ii) communication between end-users, sensors, and edge cloud; (iii) AI-based inference computation; and (iv) providing feedback to participants in the intersection. The advanced "radar-screen" application is intended to broadcast the positions and velocities of objects to intersection participants in real time.

%%%%%%%%%% Intersection map w/ cameras figure %%%%%%%%%%%%%%%%%
\begin{figure}
\centerline{\includegraphics[width=18.5pc]{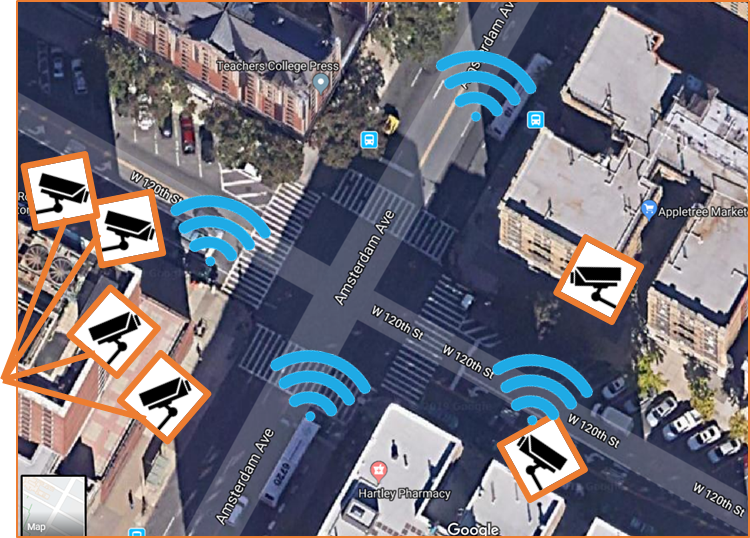}}
\caption{COSMOS pilot site with cameras and edge-cloud nodes.}
\label{COSMOS_pilot_intersection}
\end{figure}
%%%%%%%%%%%%%%%%%%%%%%%%%%%%%%%%%%%%%%%%%%%%%%%%%%%%%%%%%%%%%%%

\subsection{Privacy}
%1/4 col
% outline problem with using ground floor videos and with AI security concerns in general
% - Face/license exposure
% Alex write.

Smart-city implementations prior to 2022 indicate that privacy and data security are the key concerns impeding successful large-scale deployments. Privacy concerns are further amplified when video recordings are a part of data acquisition and processing. The COSMOS research program has a strong community outreach component. This is exemplified by multi-year activities on running NSF REM and RET programs where teachers from Harlem and other New York City schools get training and participate in developing STEM educational material for students in underprivileged schools (https://www.cosmos-lab.org/outreach/, \cite{10.1145/3431832.3431839}). 
Our approach to privacy is to integrate local communities into the data governance process. 
We will develop technologies that enable the communities to define and control data acquisition and processing supported by edge computing and temporary data storage paradigms.
%In the development of intelligent systems that rely on large amounts of visual data, privacy concerns naturally arise. How can we continue to improve technology in areas like pedestrian detection, traffic object re-identification, and tracking without compromising personal privacy? Advances in facial detection and recognition understandably inspire some Orwellian themed fears, but fortunately the same technology that enables misuse can be used to protect sensitive information. For example, our face and license plate anonymization pipeline can be used to automatically distort these areas of video feeds to prevent sensitive information from ever being recorded. 

%%%%%%%%%%%%%%%%%%%%%%%%%%%%%%%%%%%%%%%%%%%%%%%%%%%%%%%%%%%%%%%%%%%%%
%\subsection{LOW LATENCY/REAL-TIME INFERENCE - 0.25 column}
\subsection{Real-Time Interactions}
%- 0.25 column zk
The most important goal of smart city deployments is to improve the safety of pedestrians and other participants. Even in the most congested cities it is desirable to replace human drivers with safer self-driven vehicles.
This motivates the concept of cloud-connected vehicles that interact with city infrastructure to improve their ability to navigate, and requires exceptionally low closed loop latencies associated with security-critical real-time actions.  
%Mere observation of a traffic intersection provides limited benefits. The most important goal of smart city deployments should be to improve safety of pedestrians and other participants roaming the streets. Additionally, it is desirable to relive humans of having to drive vehicles even in the most congested cities. Autonomous vehicles, of the form of the year 2022, will have to evolve into cloud-connected vehicles that interact with city infrastructure to improve their ability to navigate, and potentially be "instructed" by the city traffic management how to progress.  To make it possible to increase safety and add sophisticated services, it is necessary to provide alerting and intelligent feedback within the time that makes it possible to avoid a dangerous or undesirable event, which brings about the notion or real-time. 

%%%%%%%%%%%%%%%%%%%%%%%%%%%%%%%%%%%%%%%%%%%%%%%%%%%%%%%%%%%%%%%%%%%%%
%\subsubsection{What is Real-Time in Safety-Critical Applications - 0.5 column zk}
\textbf{Real-Time for Safety-Critical Applications}
% - 0.5 column zk

%The dynamics on the streets of a metropolis are determined by vehicles moving at velocities which range from 1 to 60 miles per hour ($mph$). Extracting intelligence which indicates danger and providing feedback to vehicles or pedestrians presents computational and latency challenges. Consider the speed of $6.21~mph$ which is approximately $10~kmph$ - a vehicle maneuvering an intersection with active pedestrians should not be moving much faster. This equals to $3$ meters per second, and divided with $30$ (as in $30$ frames per second of conventional videos) yields a movement of $10$ centimeters while a single frame of video is recorded. If feedback could be provided with a latency no larger than $33.3$ milliseconds ($1$sec/$30$), so that a vehicular actuator (brake) could commence the breaking, a life might be saved. These are somewhat speculative numbers, but they are a reasonable target for exploring the architectures a smart-city intersection intelligent node. 

Extracting intelligence that indicates a potential collision and providing feedback to vehicles or pedestrians presents computational and latency challenges. City street dynamics are determined by vehicles travelling at velocities between 0 and 100 kilometers per hour ($km/h$). If we consider a vehicle travelling at $10 km/h$, a typical speed of vehicles in congested intersections, the vehicle is moving at approximately 3 meters per second ($m/s$). If we divide $3 m/s$ by the standard frame rate of conventional video, $30$ frames per second, the result is a movement of $10 cm$, or the distance travelled by the vehicle in $33.3$ milliseconds. If a vehicle's breaks could be activated in that time, it is conceivable that numerous life-threatening traffic accidents can ultimately be avoided. This approximate calculation leads us to target latencies below $33$ milliseconds.

%%%%%%%%%%%%%%%%%%%%%%%%%%%%%%%%%%%%%%%%%%%%%%%%%%%%%%%%%%%%%%%%%%%%%
%\subsubsection{Sensor Latency}
\textbf{Sensor Latencies} 

%PASSED (5)Sensor latency repeat a large portion of text from the introduction. Do we need to keep this section?
% - 0.25 column zk
Smart city sensors will have a wide range of operational frequencies and data acquisition bandwidths. $CO_2$ sensors may collect several bytes per hour, whereas high resolution cameras may stream data in compressed form at tens of Megabits per second, or in uncompressed form at several Gigabytes per second. Low-cost CMOS imaging sensors have latencies of several milliseconds, which are low enough not to obstruct the closed-loop target of 1/30 second.
%is amenable for providing services which are the target of this work, with desired closed-loop latency of 1/30 second.
IP cameras use video encoding and streaming protocols that, because of inter-frame coding, may have buffers requiring hundreds of milliseconds to decode; this process severely impedes the ability to provide closed-loop services with less than 33.3 ms latencies.

%%%%%%%%%%%%%%%%%%%%%%%%%%%%%%%%%%%%%%%%%%%%%%%%%%%%%%%%%%%%%%%%%%%%%
%\subsubsection{Communications Latency }
\textbf{Communications Latencies}  
% - 0.25 column 

Communications and networking latencies are determined as much by speed of physical media as they are driven by protocols at the application layer. The COSMOS optical network can provide up to 100 Gb/s, offering almost unlimited raw speed. On the other hand, conventional streaming of high resolution videos can create hundreds of milliseconds of latency. This suggests that video processing and inference is best done at the "extreme" edge - right next to the video sensor. More interestingly, this motivates research on integrated coding and video transmission protocols optimized for ultra-low latency transmission of videos over high bandwidth edge  communications infrastructure.

%%%%%%%%%%%%%%%%%%%%%%%%%%%%%%%%%%%%%%%%%%%%%%%%%%%%%%%%%%%%%%%%%%%%%
%\subsubsection{Inference and Decision Latency - 0.25 column}
\textbf{Inference and Decision Latencies}

% - 0.25 column zk
Inference latencies come from video preprocessing and deep learning algorithms for multiple object detection and tracking. The training of DL models is done offline and does not impact latencies for real-time interactions. 
Both published work and our own studies indicate that contemporary GPUs within specialized pipelines such as NVIDIA TensorRT and DeepStream can deliver speeds above $30$ fps for object detection and tracking. 
We previously showed that inference speed varies as a function of input resolution and actual device capabilities, but we assess that inference computation will not be a bottleneck in meeting our real-time latency target. 

The decision process is defined as a higher level of intelligence built on top of object detection and tracking. For example, this process would deduce the implications of a pedestrian being on a trajectory to intersect with a speedy vehicle and create a warning (or even a command) for the pedestrian or vehicle. Computational needs for this type of processes are subject to ongoing studies, but it is expected that the latencies will be less than a millisecond.
%, since the input will be the "compressed/meta data" and the processes will have a low-pass filtering nature. 

%%%%%%%%%%%%%%%%%%%%%%%%%%%%%%%%%%%%%%%%%%%%%%%%%%%%%%%%%%
\subsection{COSMOS Experimental Testbed}

%%%%%%%%%%%%%%%%%%%%%%%%%%%%%%%%%%%%%%%%%%%%%%%%%%%%%%%%%%
%\textbf{New York City (NYC), an example of a busy metropolis}, provides unique challenges for the development and deployment of these technologies.  
%\textbf{New York City (NYC) 
New York City (NYC) is an excellent example of a busy metropolis which provides formidable challenges for the deployment of smart city technologies.
Busy urban traffic intersections have a large number of vehicles and pedestrians moving in many directions at various speeds, often with chaotic or unpredictable behavior. Furthermore, obstructions like building corners, parked vehicles, and construction equipment present difficulty to autonomous vehicle sensors requiring further advancements in traffic intersection based automation of monitoring, measuring, learning, and feedback. 
%Smart city intersections are fundamental to the AI-based traffic management systems for future metropolises. 
%With the goal of exploring how to improve pedestrian safety and support real-time collaborative control of cloud connected vehicles, we performed real world experiments. 

%%%%%%%%%%%%%%%%%%%%%%%%%%%%%%%%%%%%%%%%%%%%%%%%%%%%%%%%%%
%\textbf{COSMOS testbed}, 
The COSMOS testbed, NSF-funded Cloud Enhanced Open Software Defined Mobile Wireless Testbed for City-Scale Deployment~\cite{raychaudhuri2020challenge}, provides an experimentation platform for applications and architectures to support intelligence nodes of future metropolises. 
For our research, we use the COSMOS pilot site located at Columbia University, in New York City, at the intersection of the 120th Street and Amsterdam Avenue.
%In support of this, 
The pilot node includes two street level and two bird’s eye cameras, as illustrated in Fig.\ref{COSMOS_pilot_intersection}.
The COSMOS edge cloud servers can run real-time algorithms for detection and tracking of objects in the intersection to monitor and manage traffic flow and pedestrian safety. 
%An advanced "radar-screen" intersection application is intended to broadcast the positions and velocities of objects to intersection participants in real-time.
The node is equipped with an optical x-haul transport system that connects AI-enabled edge computing clusters.
This allows for baseband processing with massively scalable CPU and GPU resources with FPGA assist, which can also support software defined radios. Four technology layers are provided for experimentation: the user device layer, radio hardware and front-haul network resources, radio cloud, and general purpose cloud. 

%%%%%%%%%%%%%%%%%%%%%%%%%%%%%%%%%%%%%%%%%%%%%%%%%%%%%%%%%%%%%%%%%%%%%
%\section{Building Blocks of Smart City Intersections}
%\section{Building Blocks of an Intelligent Smart Intersection Node}
\section{Building Blocks of Intelligent Nodes}

%\section{TECHNOLOGICAL COMPONENTS}
% 0.25 column zk
% choose section title. Options: Cornerstone methodologies,  Buliding Blocks, Backbone

As of 2022, individual technological modules for implementing the vision of smart cities exist in the form of low power chips, high bandwidth modems, wired and wireless networks, and GPUs for machine learning (ML) and deep learning (DL). However, major challenges exist in the domains of privacy preservation, security, intelligent decision making, system integration, and in the interactions between technology and social good.

%%%%%%%%%%%%%%%%%%%%%%%%%%%%%%%%%%%%%%%%%%%%%%%%%%%%%%%%%%%%%%%%%%%%%
\subsection{Sensors}
% - 0.5 column 
%DONE IoT, lidars, cameras, multi-modal sensor integration.
Sensors range from dozens of low rate IoT-based devices collecting data about pollution to several high resolution lidars and cameras providing real-time feeds.
Multi-modal data aggregation and collaborative intelligence are research topics of notable importance to smart intersection nodes~\cite{9112216}.  

%%%%%%%%%%%%%%%%%%%%%%% Figure - four scenes from COSMOS cameras %%%%%%
% 4 pictures in one, for camera views
% DONE Alex: enter the names the the four consitutent figures here:
% DONE Alex: enter the name, location and the name of google graphs which combined the 4 figures, here: testbed_scenes.png, ?, ? 
% DONE Alex: upload the 4 constituent figures into the "image" directory in this overleaf project
\begin{figure}
\centerline{\includegraphics[width=18.5pc]{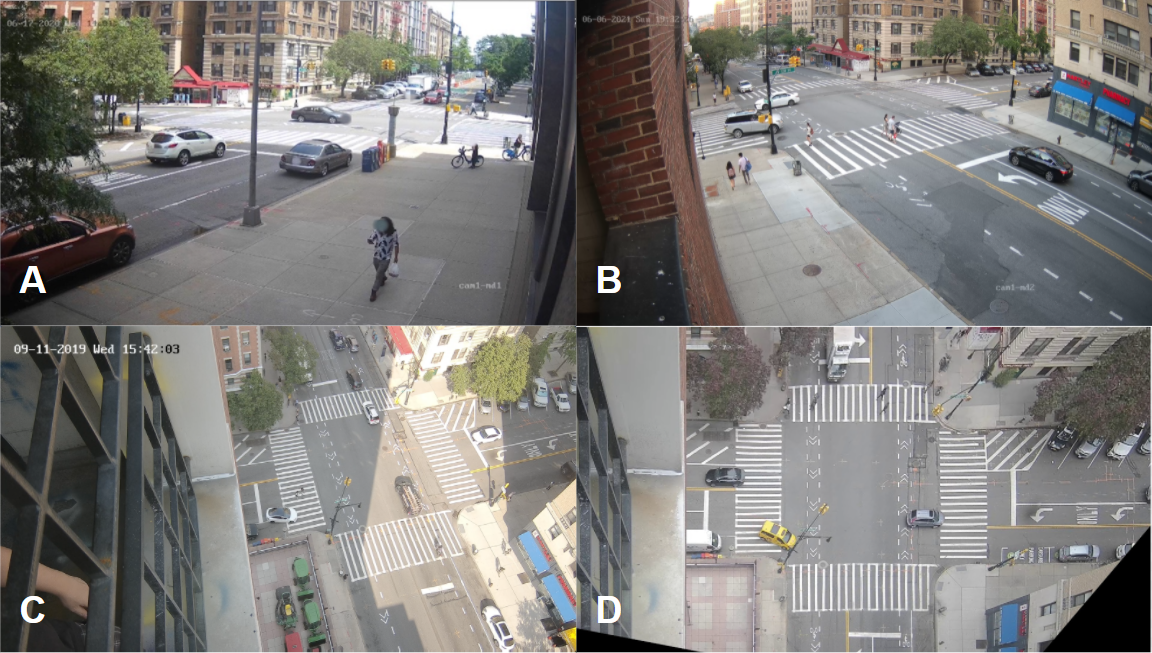}}
\caption{COSMOS testbed camera views: (A) 1st-floor camera, 120th St; (B) 2nd-floor camera, Amsterdam Ave.; (C) 12th-floor camera, Amsterdam Ave.; (D) Calibrated 12th-floor camera.}
\label{testbed_scenes}
\end{figure}
%%%%%%%%%%%%%%%%%%%%%%%%%%%%%%%%%%%%%%%%%%%%%%%%%%%%%%%%%%%%%%%%%%%%

%%%%%%%%%%%%%%%%%%%%%%%%%%%%%%%%%%%%%%%%%%%%%%%%%%%%%%%%%%%%%%%%%%%%%
\subsection{Networking}
% - 0.5 column zk
%High speed low latency networking is essential for support of real-time AI-based smart city intersection applications. 
For high bandwidth applications, networking at one intersection has to support wireless and wired connectivity from half a dozen infrastructure-installed cameras. Whereas coded video from a conventional IP-camera may require sub-hundred Mb/sec, experimentation with ultra low latency provides motivation to send raw video at several Gb/sec per camera. Support for cloud-connected vehicles could require harvesting videos and other data from each vehicle wirelessly, in either raw or meta format. 
%There may be dozens of vehicles in an intersection. 
Conventional video streaming protocols may be inadequate for accomplishing very low latencies, so research into edge-streaming protocols is an appealing topic.   
%DONE is there a references, or is more discussion needed? no

%%%%%%%%%%%%%%%%%%%%%%%%%%%%%%%%%%%%%%%%%%%%%%%%%%%%%%%%%%%%%%%%%%%%%
\subsection{Edge Computing}
%\subsection{EDGE COMPUTING}
% - 0.5 column zk

Smart city intersection applications require substantial computational resources, demand minimal latencies, and their functionality can be constrained to a limited geographical area. Furthermore, data privacy, security, and local data governance are of utmost importance. 
This strongly implicates edge computing as the right modality.
Two forms of edge computing can be used. In the extreme, AI-based computing can be done on devices located at the sensors such as Nvidia Jetson Nanos or ML-enabled ARM M1-M4 processors integrated into IoT chips. 
%(DONE consider citation for Jetson Nano) - No 
%(DONE consider citation,ask Fred) - No
On the other hand, a more powerful computing node can be located in a facilities room of a building at the intersection. The node is then connected to sensors by high speed wireless, wired, or optical infrastructure. To support low latencies from sensors to actuators via AI computing, an edge computing node has to be integrated tightly with the network communications infrastructure.

%%%%%%%%%%%%%%%%%%%%%%%%%%%%%%%%%%%%%%%%%%%%%%%%%%%%%%%%%%%%%%%%%%%%
\subsection{AI-Enabled Data Processing}
%- 0.5 column  zk
Intelligent tasks supporting smart city intersections are varied in complexity: $CO_2$ sensors generate several bytes once per hour,
whereas high resolution cameras in our studies generate Megabits per second
to be analyzed by visual deep learning models for object detection, tracking, and intelligent decisions for actuators. 
Automation and AI are crucial to scale systems for highly congested traffic intersections. 
Off the shelf AI models must be modified and retrained to accommodate the peculiarities of smart city intersection applications - one example being the detection of tiny pedestrians when viewed from bird's eye cameras. 
%Edge computing nodes have to be equipped with real-time AI functionality that can provide high accuracy and ultra-safe inference. 

%%%%%%%%%%%%%%%%%%%%%%%%%%%%%%%%%%%%%%%%%%%%%%%%%%%%%%%%%%%%%%%%%%%%%
%\subsubsection{Data Preprocessing}
\textbf{Data Preprocessing}

% Zhengye and/or Alex write.
% 1/2 col
% - video data from multiple views - combining these different views
% - Figure: 1st + 12th Floor views figure from SD paper
% - 12th floor calibration
% - annotation/labeling
% - Figure: calibration figure from resolution paper
% - Background subtraction/blocking out of irrelevant areas
% - Pretraining + Data augmentation
%To create effective deep learning based tools, 
Visual deep learning tools require 
data preparation, labeling, and augmentation. 
The COSMOS pilot node
%, at 120th Street and Amsterdam Avenue, we have 
contains low-elevation cameras and high-elevation bird’s eye view cameras, each requiring different type of preprocessing  
%(see Figure~\ref{ground_birdeye_view}). 
(Fig.~\ref{testbed_scenes}).
The variation in angles and distances to the intersection, scale of objects, and overlapping field-of-views 
allow experimentation with the best view for a given application. For example, ground floor cameras are closer to traffic objects. They consequently provide more visual details for applications such as multi-camera object reidentification, but are not as well suited to analyze large scale traffic patterns due to the scale distortion between objects at varying distances to the camera – the bird’s eye view cameras offer a better perspective for this type of application. 

\begin{figure}[!t]
\centerline{\includegraphics[width=18.5pc]{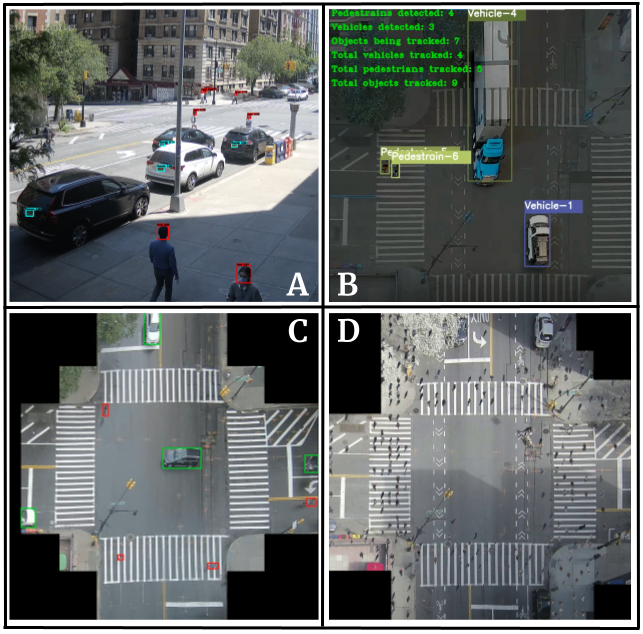}}
\caption{(A) YOLOv4 detections of faces and license plates in ground floor video; (B) SORT tracking of vehicles and pedestrians in bird's eye video; (C) Bird's eye ground truth bounding box labels of intersection objects; (D) Pedestrian copy-paste data augmentation for improving detection of small objects}
\label{application_visualizations}
\end{figure}

\begin{figure}
\centerline{\includegraphics[width=18.5pc]{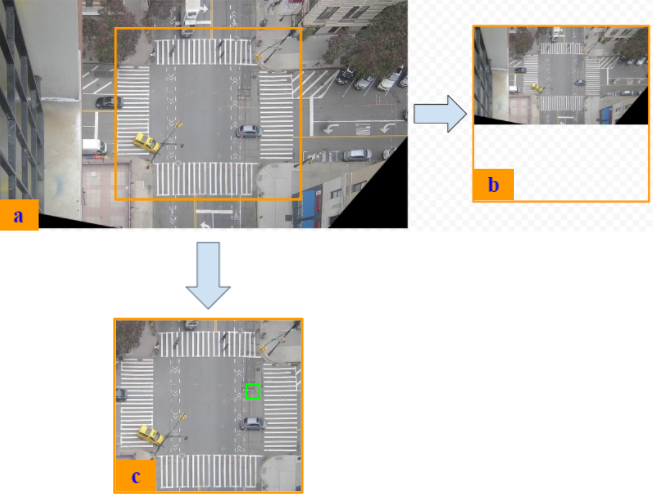}}
\caption{(a) Calibrated 16:9 native frame; (b) 16:9 frame squared using zero-padding; (c) Square cropped frame.}
\label{input_methods}
\end{figure}

High-elevation cameras allow us to perform calibration transforms to improve the effectiveness of deep learning models. 
See in 
%Figure~\ref{12thFloor_OriginCalibViews}
Fig.~\ref{testbed_scenes}
and Fig.~\ref{input_methods} that the high-elevation camera view can be adjusted to appear perpendicular to the road by applying a homography transformation, after which resizing and cropping of the frame create the square aspect ratio required by many DL models. In our traffic intersection use case there are locations in the frame where relevant objects do not appear (i.e. no cars on building walls or pedestrians flying in the air). 
This motivates the creation of (black) masks overlayed on top of the frames, as seen in 
Fig.~\ref{application_visualizations} and  Fig.~\ref{input_methods}.

Supervised object detection and tracking models require a large number of precisely annotated ground truth labels to train the algorithms “by example”. Producing accurate and consistent sets of labeled videos is difficult as both domain knowledge and significant amounts of time are needed. 
To support our experiments we annotated thousands of frames capturing the intersection in various weather, lighting, and congestion conditions.

% link to drawings directory: https://drive.google.com/drive/folders/11ox-Pb1wLmtk-YAdLplmAMpBGrUDheLB

%%%%%%%%%% Detection and Tracking figure %%%%%%%%%%%%%%%%%
%%%%%%%%%%%%%%%%%%%%%%%%%%%%%%%%%%%%%%%%%%%%%%%%%%%%%%%%%%%%%%%%%%%%%

Object detection models typically struggle with small object detection. Tiny pedestrians in the bird's eye camera view, as well as far-away license plates in the ground-floor camera view, convey very little information. This results in relatively poor detection and tracking accuracies. To improve the performance, we have deployed techniques of pretraining the DL models with a small-object dataset~\cite{zhu2018visdrone} and applying data augmentation techniques such as the copy/paste method illustrated in Fig.~\ref{application_visualizations} (D).

%%%%%%%%%%%%%%%%%%%%%%%%
\begin{table}[!t]

  \begin{threeparttable}
   
    \caption{Object Detection Performance}
     \label{tab:tabdet_perf}
     \begin{tabular}{p{1cm}p{1.cm}p{1.cm}p{1cm}p{1.cm}}
        \toprule
       \textbf{Model} & \textbf{Pedestrian AP (\%)} & \textbf{Vehicle AP (\%)} & \textbf{mAP (\%)} & \textbf{Inference Speed*}\\
        \midrule
\textbf{YOLOv4} & \textbf{66.31} & \textbf{97.58} & \textbf{81.95} & \textbf{34.99} \\ 
\textbf{SSD} & 57.04 & 94.81 & 75.93 & 11.31 \\
\textbf{RetinaNet} & 20.83 & 95.59 & 58.21 & 22.97 \\ 
        \bottomrule
     \end{tabular}
    \begin{tablenotes}
      \small
      \item *Inference speed (FPS) on NVIDIA T4 GPU.
    \end{tablenotes}
  \end{threeparttable}

\end{table}
%%%%%%%%%%%%%%%%%%%%%%%%%%%

%%%%%%%%%%%%%%%%%%%%%%%%%%%%%%%%%%%%%%%%%%%%%%%%%%%%%%%%%%%%%%%%%%%%%
%\subsubsection{OBJECT DETECTION AND TRACKING}
%\subsubsection{Object Detection and Tracking}
\textbf{Object Detection and Tracking}
%  -1.5 column
% - DONE Table: model comparison study (2022)

\begin{figure}
\centerline{\includegraphics[width=18.5pc]{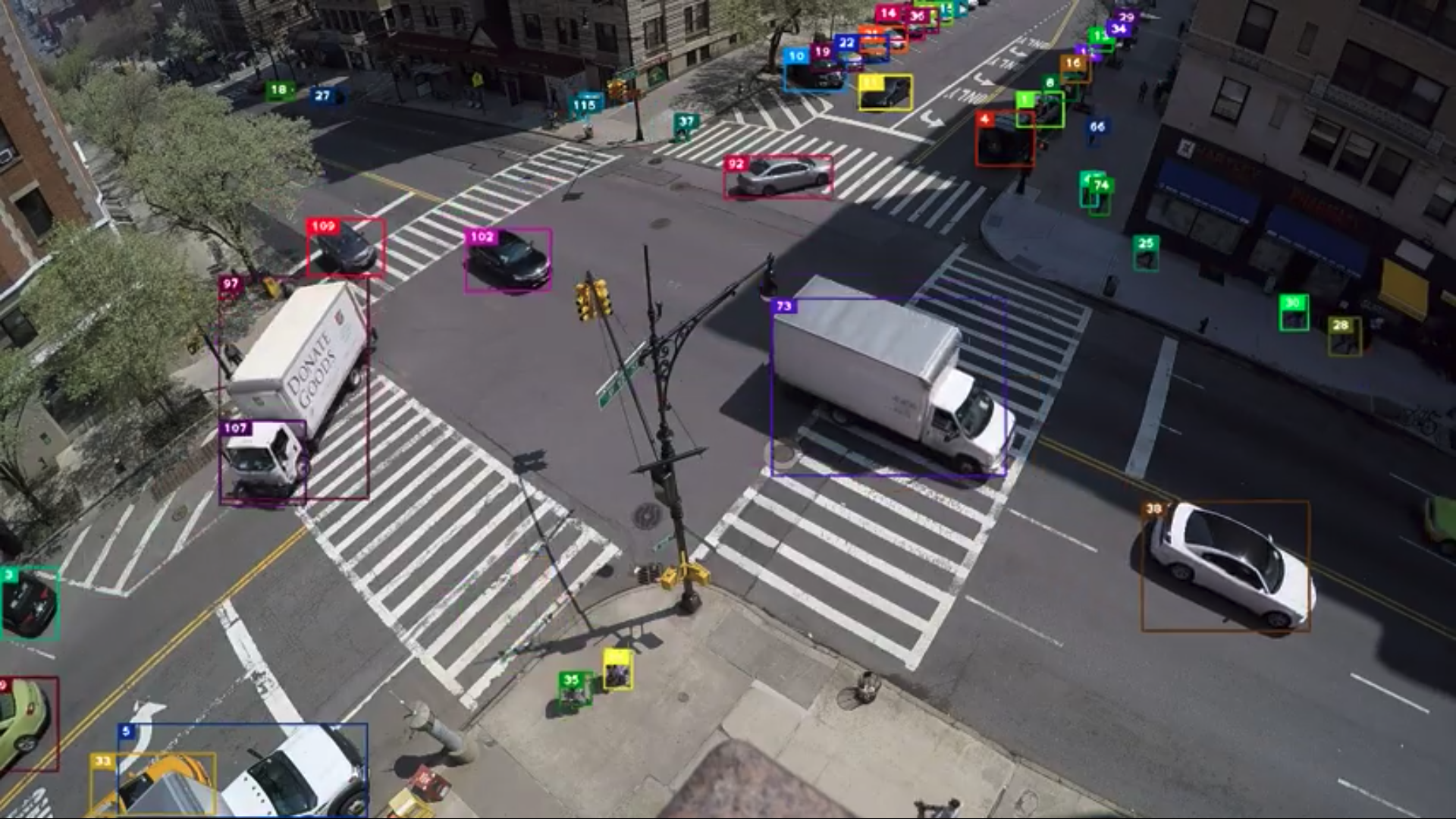}}
\caption{Pedestrian and Vehicle detection on 120th St. and Amsterdam Ave.}
\label{detection_preview}
\end{figure}

In smart traffic intersections, detecting pedestrians and vehicles and tracking their trajectories are the prerequisites for all downstream applications, Fig.  \ref{detection_preview}. 
This involves two computer vision tasks: 
Object Detection and Multiple Object Tracking (MOT). 
%DONE where should be the first time we define MOT, should it be earlier when we discuss tracking?
The objective of object detection is to localize and classify objects within the frame. MOT aims to associate object identities across successive frames. 
State-of-the-art methods rely on deep learning blocks such as Convolutional Neural Networks (CNN)\cite{liu2020deep} and Vision Transformers~\cite{dosovitskiy2020image}. 
These methods bring heavy computational cost, 
and the accuracy-speed trade-off - the budgeting between computational complexity and inference speed - is vital to the success of smart city applications. With this consideration in mind, we experimented with a series of algorithms for detecting and tracking objects to find the best approach~\cite{yang2020cosmos} 
%es in smart city application
based on our custom annotated dataset for bird's eye videos. 
We choose YOLOv4~\cite{2020arXiv200410934B} as the base detector for all downstream applications since it is able to provide accurate results in real-time. 
Object detection performance is shown in Table~\ref{tab:tabdet_perf}, where the Average Precision (AP) and mean Average precision (mAP) are used as the evaluation metrics. On our bird's eye view intersection data, YOLOv4 outperforms both RetinaNet~\cite{lin2017focal} and SSD~\cite{liu2016ssd} in terms of AP and inference speed, where inference speed is measured as the average time for a forward pass through the model with batch size equal to 1. 
For MOT, different scenarios need to be considered separately. For bird's eye cameras, 
object occlusions barely occur, so re-identification (reID) calculation is not as necessary as for the ground level cameras. The reID calculation is often the computation bottleneck in MOT algorithms. "Simple Online and Realtime Tracking" (SORT) and "Simple Online and Realtime Tracking with a Deep Association Metric" (DeepSORT) suffice for the bird's eye view cameras.
Illustrations for detection are shown in Fig.~\ref{application_visualizations}.

%%%%%%%%%% figure 4th Floor Detection example %%%%%%%%%%%%%%%%%
%%%%%%%%%%%%%%%%%%%%%%%%%%%%%%%%%%%%%%%%%%%%%%%%%%%%%%%%%%%%%%%

\textbf{Image Resolution and Object Density}

Highly elevated bird's eye cameras have a good view of the overall scene,  shown in
%(Fig.~\ref{12thFloor_OriginCalibViews}), old figure
Fig.~\ref{testbed_scenes}.
Pedestrians, which appear small, 
become a problem for object detection and tracking. Intuitively, the higher the resolution of the input image, the more object features can be preserved.
However, higher resolution leads to a larger computational cost, thus making the inference slow. We tested a dozen combinations of image input resolutions and aspect ratios to find the best balance between accuracy and speed, three of which are shown in Fig.~\ref{input_methods}. Some deep learning models, like YOLOv4~\cite{2020arXiv200410934B}, perform better on input images with a fixed-sized, square aspect ratio. To maximize the preservation of important features of the intersection scene and to minimize the irrelevant components, the experiments indicate that the "squared cropped" $832\times832$ input produces the best results~\cite{duan2021smart}.

%, shown in Fig.~\ref{input_methods} and Table.~\ref{table:resolutions}. Note that groups a, b, and c in the table correspond to the input methods shown in Fig.\ref{input_methods} (a), (b), and (b), respectively.
%%%%%%%%%%%%%%%%%%%%%%%%
%%%%%%%%%%%%%%%%%%%%%%%%

%%%%%%%%%%%%%%%%%%%%%%%%
\begin{figure}
\centerline{\includegraphics[width=18.5pc]{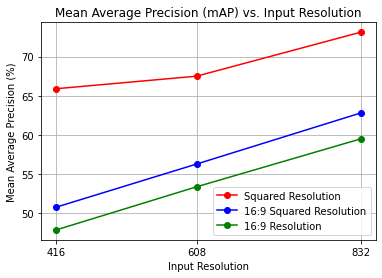}}
%\caption{Average Precision evaluation of pedestrians and vehicles for the 9 groups listed in table.\ref{table:resolutions}~\cite{duan2021smart}.}
%\caption{mAP for pedestrians and vehicles for 9 groups in Table ~\ref{table:resolutions}~\cite{duan2021smart}.}
\caption{mAP for pedestrians and vehicles, 9 cases of image resolution vs. aspect ratio} 
\label{resolution_evaluation}
\end{figure}
%%%%%%%%%%%%%%%%%%%%%%%%

Object density refers to the number of objects in a scene, which may impact the speed of inference as the busyness of the streets change through the day. We explored the inference time for ten 90-second videos where  the number of objects varied from $4,000$ to $26,000$. The results show approximately $40$ percent increase in computational load from the lowest to the highest density case. 
This is important in that it shows that object density can be used to switch between computational resources to obtain the optimal power/accuracy balance.

\section{Applications}

Advances in video based object detection and tracking have enabled the deployment of a number of traffic intersection applications, where one can identify the locations of objects in the intersection and classify them by type of vehicle, pedestrian, bicycle, etc. They can be tracked as unique entities which persist through the duration of traffic cycles, different camera views, and times of the day, week, month, and year. The abundance of spatial, temporal, and visual data makes it possible to perform data anonymization, quantization of traffic trends, crowd behavior surveillance, real time intersection radar mapping, and more. 

\subsection{Privacy Protection - Face and License Plate Anonymization}

%%%%%%%%%% Blurring Pipeline input/output figure %%%%%%%%%%%%%%%%%
\begin{figure}
\centerline{\includegraphics[width=18.5pc]{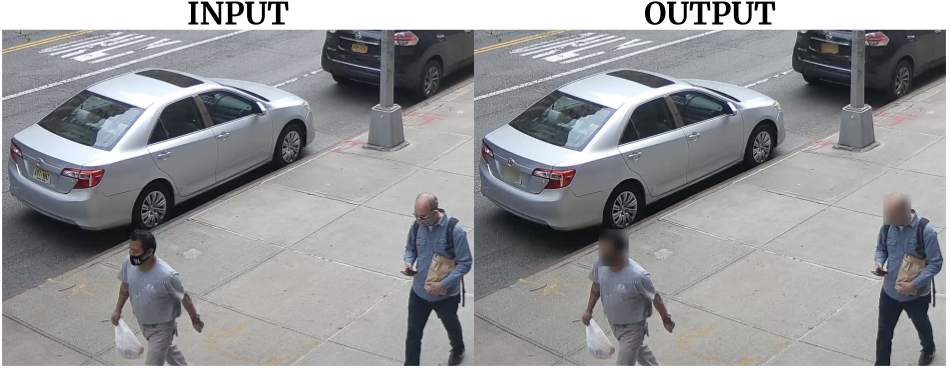}}
\caption{Input (left) and output (right) of the face and license plate blurring pipeline.}
\label{blur_sample}
\end{figure}
%%%%%%%%%%%%%%%%%%%%%%%%%%%%%%%%%%%%%%%%%%%%%%%%%%%%%%%%%%%%%%%%%%%%%

Collecting real-time images and videos of public spaces from street level inadvertently involves capturing sensitive information such as faces and license plates. To avoid leaking private information with our datasets, we generated a pipeline to automatically blur these sensitive areas. We trained several object detection models on a custom labeled dataset to detect faces and licenses for subsequent anonymization. When training with sequential video datasets, it is important to leave entire videos out of the training process to use for validation. Stationary objects – parked cars, seated pedestrians, chained bicycles, etc. – occur identically in many frames, and model evaluation on these stationary objects yields biased results. This leads to model overfitting and poor generalization to new intersection scenes, which has to be addressed.

Fig.~\ref{blur_sample} shows an example input and output frame of the anonymization pipeline. For our face and license detection model, we chose YOLOv4 ~\cite{2020arXiv200410934B} for its compromise between detection accuracy and inference speed. 
%As summarized in Table~\ref{tab:tabdet_perf}, we are able to achieve high mAP with various model input resolutions. 
For privacy critical applications, the most relevant performance measure is recall, the number of relevant faces and licenses that are detected out of the total number that pass through the frame. False positives are less of an issue than false negatives, as they result in an extra blurred area of the frame, but not a privacy leak. In our case, not all faces and licenses are “relevant” – some are too far away and too low resolution to be identifiable. We exclude these instances from the recall evaluation by defining pixel area thresholds below which the objects are ignored. We found that, below certain thresholds, facial features and license plate characters could not be reliably identified. While there exist information reconstruction techniques that could potentially recover these features,  %features~\cite{liang2021swinir}, 
%DONE zoran consider removing this reference, to have space for some other more important reference - replaced by our paper Smart City Traffic Intersection: Impact of Video Quality and Scene Complexity on Precision and Inference
this is outside the scope of this project to consider them. Furthermore, we would need to reconsider our choice of anonymization as any form of blurring becomes ineffective. In the visible object evaluation our pipeline blurs over 99\% of visible faces and licenses and in the total evaluation it blurs over 96\% of objects greater than 100 pixels. 

\begin{figure}
\centerline{\includegraphics[width=18.5pc]{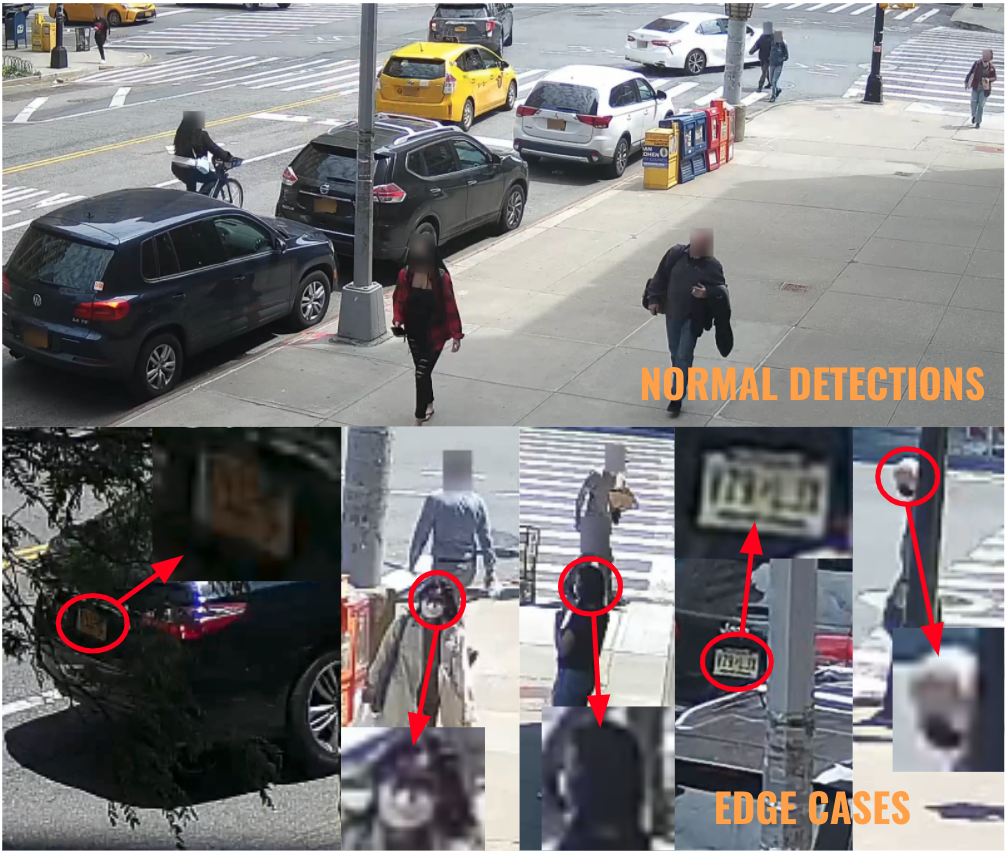}}
\caption{Normal blurring pipeline detections (top) vs. edge cases (bottom).}
\label{blurEdgeCases}
\end{figure}

To increase our confidence in the anonymization pipeline, we performed manual evaluations by inspecting anonymized output videos for misses, where a miss is defined as an object with more than a quarter of the face or license plate exposed. The results of the manual evaluations confirmed the results of the programmatic evaluations and shed some light on edge cases where our models consistently missed, 
Fig.~\ref{blurEdgeCases}. Most edge cases were due to occlusions such as occluded borders of license plates, pedestrian body occlusion, and tree branch occlusion, resulting in consistent false negatives. More data collection and training is needed to rectify these edge cases. 

%%%%%%%%%% Blurring Edge cases figure %%%%%%%%%%%%%%%%%
%%%%%%%%%%%%%%%%%%%%%%%%%%%%%%%%%%%%%%%%%%%%%%%%%%%%%%%%%%%%%%%%%%%%%

%%%%%%%%%%%%%%%%%%%%%%%%%%%%%%%%%%%%%%%%%%%%%%%%%%%%%%%%%%%%%%%%%%%%%
%\subsection{COUNTING OBJECTS FOR TRAFFIC MANAGEMENT - 1 column}
%\subsection{Counting Objects for Traffic Management - 1 column}
\subsection{Counting Objects}

An important goal for smart intersections is to analyze traffic flow in real time. To this end, we use detection and tracking to classify and count vehicles and pedestrians and follow their paths through the intersection. Accumulation of the tracks provides sufficient data for traffic trend analyses that can be used to optimize traffic flow and improve pedestrian safety in the intersection. 

To perform object tracking we use the detection based (MOT) algorithm DeepSORT. DeepSORT requires an object detection model to provide the locations and features of an object to be tracked. Given detections of vehicles and pedestrians, DeepSORT uses a Kalman filter to map detections with similar sizes and motions across frames of a video. In this way we can assign IDs to detected objects that persist throughout multiple video frames. Additionally, DeepSORT uses visual features of the object to increase the reliability of the tracking. Even if the object fails to be detected in consecutive frames, it can be assigned to the correct track by the re-identification model (reID) based on its visual features. 

Though DeepSORT is a robust tracking system, it is still dependent on high quality object detection. If an object is not detected or misclassified for multiple consecutive frames, it will be regarded by the algorithm as a “new track” – the old track disappears and a new one is created upon redetection. For vehicles, we achieve consistent high accuracy detection, and corresponding high accuracy tracking, but for pedestrians, which have 4-5x smaller cross sections, high accuracy detection is a more difficult task. Pedestrian tracking accuracy suffers as a result of lower accuracy pedestrian detection. Data augmentation techniques such as the copy-paste pedestrian method shown in Fig.~\ref{application_visualizations} (D) and pretraining object detectors on small-object datasets show improvement for small-object detection, but pedestrian detection and tracking accuracies are still lower than for vehicles, with multiple object tracking accuracies (MOTA) of 75.16\% and 18.23\% for vehicles and pedestrians, respectively. 

The vehicle tracking performance is sufficient for applications that quantify traffic flow. For example, in an automatic counting task we record vehicles passing through the intersection as turning right, turning left, or going straight from all four directions with an accuracy of 95\% evaluated over 21 minutes of video recording.

\subsection{Social Distancing in Pandemics}

Smart cities can assist in combating global pandemics, such as COVID-19, by providing means for monitoring, analyzing, and potentially controlling social distancing behavior. We proposed several techniques and applied them to video datasets collected at the COSMOS pilot intersection. 

The fundamental idea is to estimate distances between pedestrians and compare them against the recommended minimal distance threshold. The first step is to detect the pedestrians. The real-world distance is then estimated by calculating the pixel-wise distance between pedestrians within one frame. The tracking of pedestrians between frames facilitates the calculation of higher order statistics, related to safe social distancing groups, which are more meaningful than an individual-to-individual social distancing violation rates. When acquaintances are walking together on the street as a "safe group", the intra-group distance is often smaller than the social distancing  threshold, which triggers the indication of the violation. To solve this problem, we utilize the pedestrian trajectory similarity and stability, which can evaluate the motion dynamic between every pedestrian pair. This group validation approach is able to significantly reduce the number of false positive violations, achieving the F1 score of $0.92$. Based on this approach, we built a social distancing analysis system B-SDA~\cite{yang2021birds} for bird's eye view cameras, as well as a complementary method Auto-SDA~\cite{ghasemiyangdemo,ghasemi2021auto} with ground level cameras.

%%%%%%%%%%%%%%%%%%%%%%%%%%%
\begin{figure}
\centerline{\includegraphics[width=0.99\linewidth]{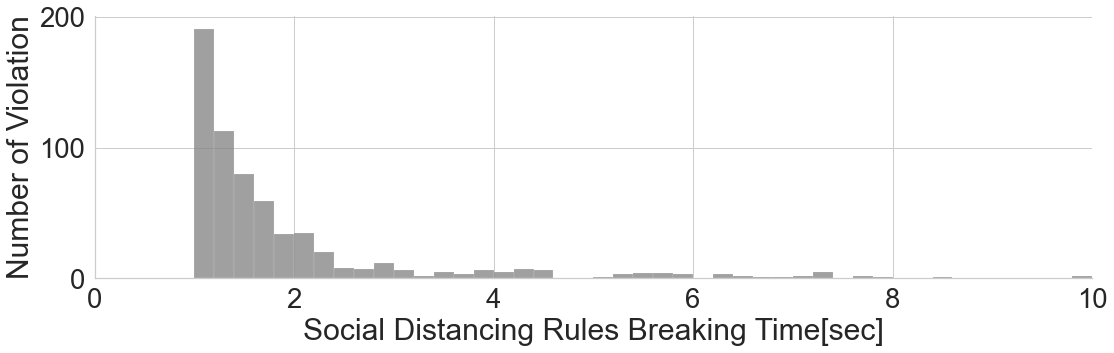}}
\caption{B-SDA: Distribution of the duration of social distancing violations.}
\label{Social_breaking_bar}
\end{figure}
%%%%%%%%%%%%%%%%%%%%%%%%%%%

\begin{figure}[!t]
\begin{center}
\includegraphics[width=0.99\linewidth]{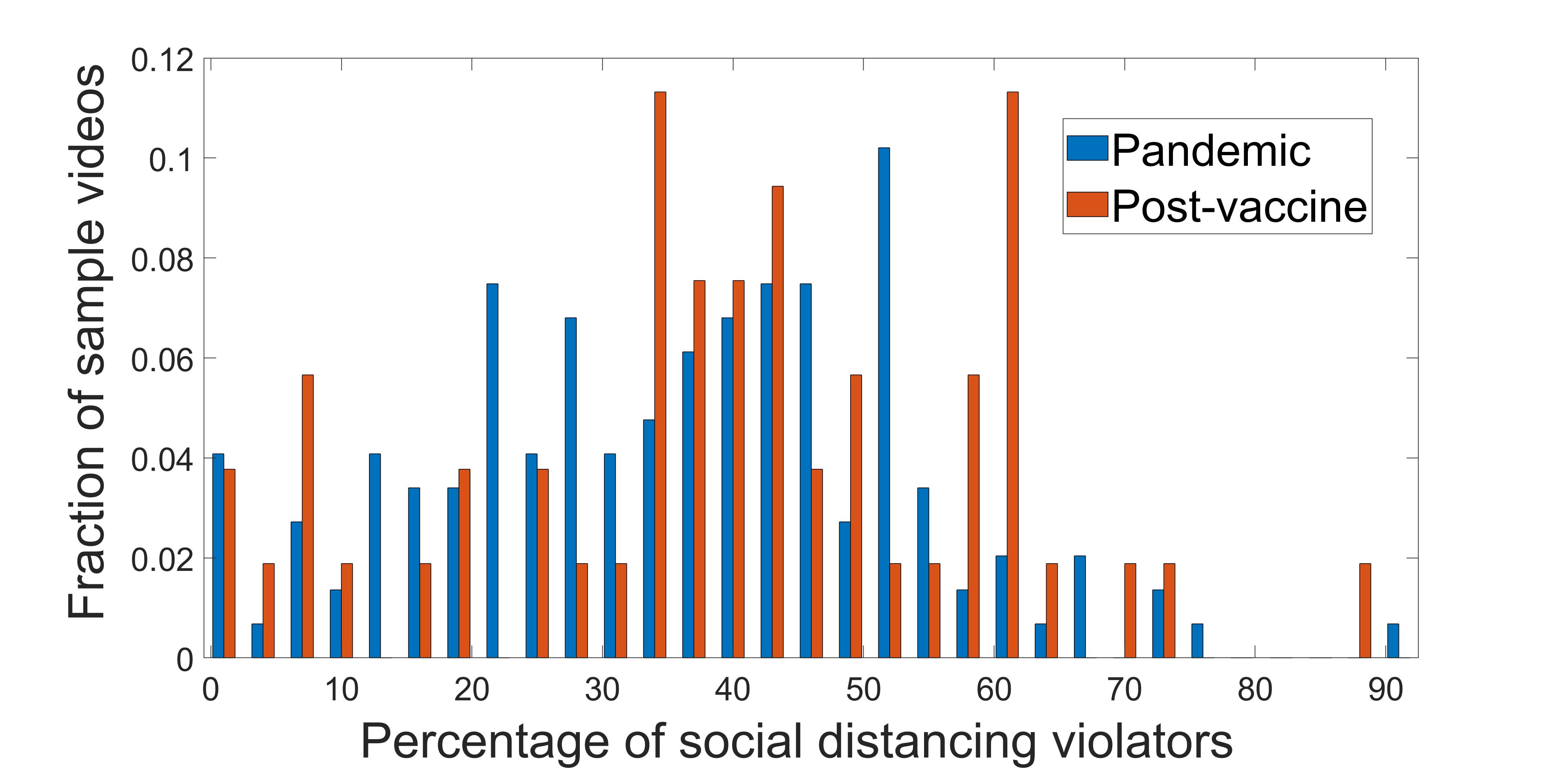}
\end{center}
\vspace*{-0.3cm}
%\caption{ Auto-SDA: Normalized histogram of the percentage of social distancing violations. ~\cite{ghasemiyangdemo}.}
\caption{ Auto-SDA: Normalized histogram of the percentage of social distancing violations.}
\label{asda_violation_comparison}
\end{figure}

An example of the results obtained with the bird's eye video dataset, illustrated in Fig.~\ref{Social_breaking_bar}, shows the distribution of the duration of social distancing violations during the Covid-19 pandemic.  Fig.~\ref{asda_violation_comparison}
shows the social distancing violation rates for the ground-floor camera dataset (i) during the pandemic and (ii) after the vaccine is widely available.
%shows the social distancing violation rates during the pandemic and after the vaccine is widely available 
Detailed analyses and comparisons of multiple statistics before the pandemic and during the pandemic demonstrate that the proposed systems can reliably identify social distancing violations.

%%%%%%%%%%%%%%%%%%%%%%%%%%
%%%%%%%%%%%%%%%%%%%%%%%%%%

\subsection{Real Time "Radar-Screen"}

\begin{figure}
\centerline{\includegraphics[width=18.5pc]{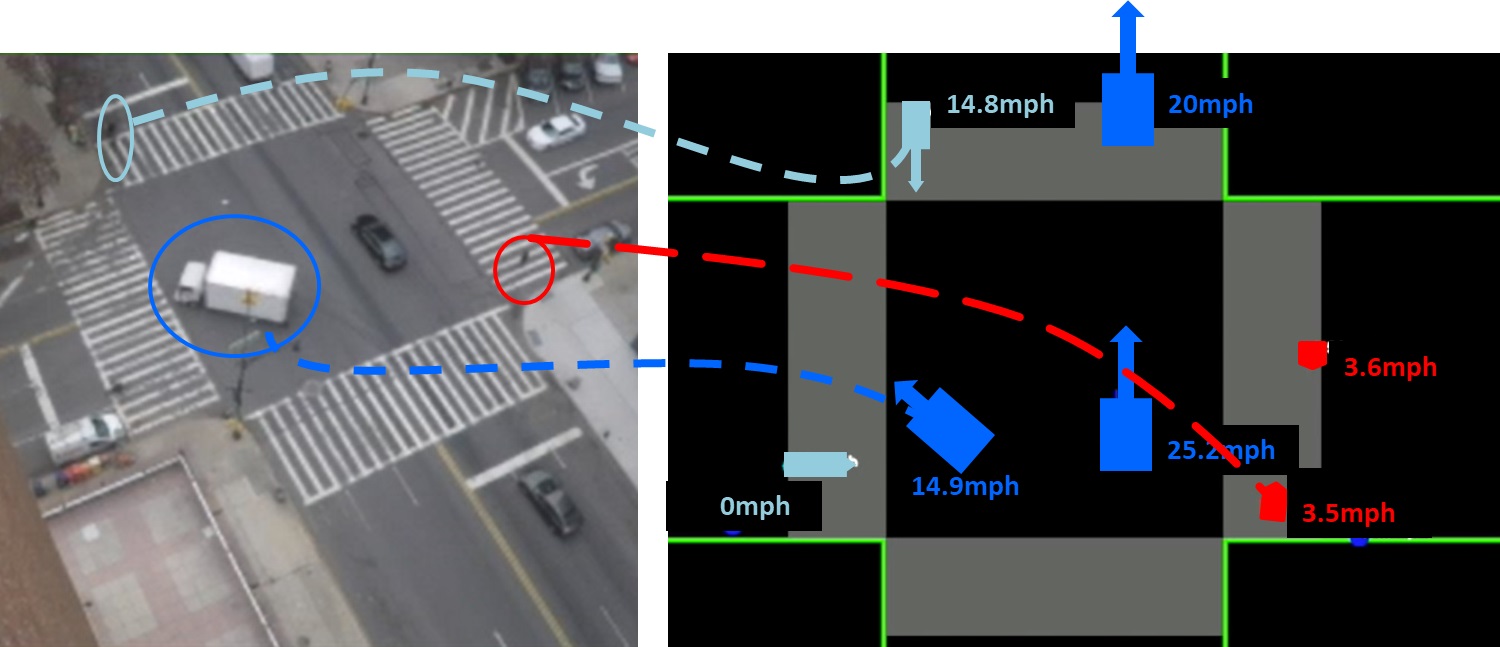}}
\caption{The “radar screen”: one frame of a video containing locations and velocities of objects within an intersection.}
\label{radar-screen-figure}
\end{figure}

The "Radar-screen" application aims to infer positions and velocities of objects within a traffic intersection and broadcast them to the participants in the intersection, as illustrated in Fig.~\ref{radar-screen-figure}. 
%We define a "radar-screen" application with the goal of inferring positions and velocities of objects within a traffic intersection and broadcasting them to the participants in the intersection, illustrated in Fig.~\ref{radar-screen-figure}. 
The information can be distributed in raw or coded/meta format.
The application intends to provide a real-time service with latency of $1/30$ seconds between the observation of objects and the wireless broadcast delivery. As described previously, this is motivated by the approximation of a $10$ centimeter vehicle movement with speed of $10$ km/h. 

The application includes the acquisition of videos from surrounding buildings, potential harvesting of videos (or encoded data) from cameras within vehicles, harvesting of IoT sensor data, transmission via a high speed network to the inference computer, data aggregation and preprocessing, DL-based object detection and tracking, extraction of information at a higher abstraction level, and (in a more advanced version) deduction of commands that may be issued to individual vehicles after optimizing the traffic flow. The final step is the broadcasting of information.
%%%%%%%%%%%%%%%%%
%%%%%%%%%%%%%%%%%%%%%%%%%%%%%%%%%%%%%%%%%%%%%%%%%%%%%%%%%%%%%%%%%%%%%
This is an aspirational application in that achieving the cumulative latency of $33.3$ ms is technologically challenging. Balancing between computational capabilities, power consumption, and latency minimization of the extreme edge compute units, or edge computing centers, requires rapid sensor data acquisition and dynamic network and resource control. This application motivates research to optimize each of the building blocks described in previous sections of this paper as well latency-focused cross-module system integration. 

\subsection{Traffic Management}

Intelligent nodes located at individual intersections provide powerful data acquisition and intelligent edge-computing. On a larger scale, smart cities require the aggregation of data from multiple intersections and mutual coordination. In that vein, we have commenced collaborative studies with traffic engineering experts on the definition of key parameters such as timing resolution, sensor locations, and APIs for data exchange between intelligent smart intersection nodes and traffic optimization systems\cite{karagiannis2011vehicular}. 
We are building simulators and defining digital twins that will play predictive roles in the behavior of individual traffic participants and in global optimization of traffic management. 

\section{Conclusion and Future Challenges}

A vision of the smart city intersection as the intelligence node for future metropolises has been presented. The proposed architecture is driven by societal needs to preserve privacy, which strongly implicate edge computing and intelligence as the key paradigm for data management and processing. Key technological components have been reviewed such as sensors, networks, and edge AI computing. Real time needs of future safety-critical systems have been examined, and design considerations for a "radar-screen" application, which closes the loop from sensors to actuators, have been summarized. The requirements for low latency, based on the 33.3 ms target, have been explored. System integration challenges have been illustrated using the examples from experiments performed on the pilot node of the COSMOS testbed in New York City. 

Our research points to the following exploration topics:  
(i) State of the art DL-based object detection models are comprised of over 60 million parameters and require passing more than 100 convolutional layers, where each convolution has $O(n^4)$ complexity. Model optimization techniques like weight pruning, inference scheduling, and neural algorithmic search strategies~\cite{banbury2020benchmarking} need to be incorporated into practical systems; 
(ii) Reliance on supervised datasets for video processing is not scalable due to the labeling cost and quality concerns. This necessitates research on unsupervised learning methodologies which should be based on continuous or active learning, and take advantage of the peculiarities of the fixed scene within a traffic intersection \cite{DBLP:journals/corr/abs-2103-11568};
% DONE alex add one reference from Chengbo
(iii) Data fusion from multiple cameras is expected to yield notable improvements in detection and tracking accuracies; 
(iv) Achieving low latency for low rate little-data applications is possible by using processing on the "extreme edge", but meeting the requirements of 1/30 second latency for high resolution videos is a challenge. New video coding methods and streaming protocols should be explored with focus on localized low-latency performance. 

\section{Acknowledgment}
This work was supported in part by NSF grants 
CNS-1827923, OAC-2029295,  CNS-2038984, CNS-1910757, and AT\&T VURI award. 

\bibliography{main.bbl}{}
\bibliographystyle{ieeetr}

%%%%%%%%%%%%%%%%%%%%%%%%%%%%%%%%%%%%%%%%%%%%%%%%%%%%%%%%%%%%%%%%%%%%%%%%%%%%%%
\end{document}